# Sample-efficient Nonstationary Policy Evaluation for Contextual Bandits

**Miroslav Dudík**[*]
Microsoft Research
New York, NY

**Dumitru Erhan**
Yahoo! Labs
Sunnyvale, CA

**John Langford**[*]
Microsoft Research
New York, NY

**Lihong Li**[*]
Microsoft Research
Redmond, WA

## Abstract

We present and prove properties of a new offline policy evaluator for an exploration learning setting which is superior to previous evaluators. In particular, it simultaneously and correctly incorporates techniques from importance weighting, doubly robust evaluation, and nonstationary policy evaluation approaches. In addition, our approach allows generating longer histories by careful control of a bias-variance tradeoff, and further decreases variance by incorporating information about randomness of the target policy. Empirical evidence from synthetic and real-world exploration learning problems shows the new evaluator successfully unifies previous approaches and uses information an order of magnitude more efficiently.

## 1 Introduction

We are interested in the "contextual bandit" setting, where on each round:

1. A vector of features (or "context") $x \in \mathcal{X}$ is revealed.
2. An action (or arm) $a$ is chosen from a given set $\mathcal{A}$.
3. A reward $r \in [0, 1]$ for the action $a$ is revealed, but the rewards of other actions are not. In general, the reward may depend stochastically on $x$ and $a$.

This setting is extremely natural, because we commonly make decisions based on some contextual information and get feedback about that decision, but not about other decisions. Prominent examples are the ad display problems at Internet advertising engines [15, 7], content recommendation on Web portals [18], as well as adaptive medical treatments. Despite a similar need for exploration, this setting is notably simpler than full reinforcement learning [25], because there is no temporal credit assignment problem—each reward depends on the current context and action only, not on previous ones.

The goal in this setting is to develop a good *policy* for choosing actions. In this paper, we are mainly concerned with *nonstationary* policies which map the current context and a history of past rounds to an action (or a distribution of actions). Note that as a special case we cover also *stationary* policies, whose actions depend on the currently observed context alone. While stationary policies are often sufficient in supervised learning, in situations with partial feedback, the most successful policies need to remember the past, i.e., they are nonstationary.

The gold standard of performance here is deploying a policy and seeing how well it actually performs. This standard can be very expensive in industrial settings and often impossible in academic settings. These observations motivate us to construct methods for scoring policies *offline* using recorded information from previously deployed policies. As a result, we can dramatically reduce the cost of deploying a new policy while at the same time rapidly speeding up the development phase through the use of benchmarks, similar to supervised learning (e.g., the UCI repository [1]) and *off-policy* reinforcement learning [21]. We would like our method to be as general and rigorous as possible, so that it could be applied to a broad range of policies.

An offline policy evaluator is usually only useful when the future behaves like the past, so we make an IID assumption: the contexts are drawn IID from an unknown distribution $D(x)$, and the conditional distribution of rewards $D(r|x, a)$ does not change over time (but is unknown). In our intended applications like medical treatments or Internet advertising, this is a reasonable assumption.

Below, we identify a few desiderata for an offline evaluator:

- *Low estimation error.* This is the first and foremost desideratum. The error typically comes from two sources—bias (due to covariate shift and/or insufficient expressivity) and variance (insufficient number of samples). Successful methods allow optimization of the tradeoff between these two components.
- *Incorporation of a prior reward estimator.* The evaluator should be able to take advantage of a reasonable reward estimator whenever it is available.
- *Incorporation of "scavenged exploration."* The evaluator should be able to take advantage of *ad hoc* past deployment data, with the quality of such data determining the bias introduced.

---
[*]This work was done while MD, JL, and LL were at Yahoo! Research.

**Existing Methods.** Several prior evaluators have been proposed. We are going to improve on all of them.

- The direct method (DM) first builds a reward estimator $\hat{r}(x,a)$ from logged data that predicts the average reward of choosing action $a$ in context $x$, and then evaluates a policy against the estimator. This straightforward approach is flexible enough to evaluate any policy, but its evaluation quality relies critically on the accuracy of the reward estimator. In practice, learning a highly accurate reward estimator is extremely challenging, rendering high bias in the evaluation results.
- Inverse Propensity Scoring (IPS) [11] and importance weighting require the log of past deployment that includes for each action the probability $p$ with which it was chosen. The expected reward of policy $\pi$ is estimated by $\frac{rI(\pi(x)=a)}{p}$, where $I(\cdot)$ is the indicator function. This formula comes up in many contexts and is built into several algorithms for learning in this setting such as EXP4 [3]. The IPS evaluator can take advantage of scavenged exploration through replacing $p$ by an estimator $\hat{p}$ [16, 24], but it does not allow evaluation of nonstationary policies and does not take advantage of a reward estimator.
- Doubly Robust (DR) Policy Evaluation [6, 23, 22, 20, 13, 7, 8] incorporates a (possibly biased) reward estimator in the IPS approach according to:

$$(r - \hat{r}(x,a))I(\pi(x) = a)/p + \hat{r}(x, \pi(x)) \quad , \quad (1.1)$$

where $\hat{r}(x,a)$ is the estimator of the expected reward for context $x$ and action $a$. The DR evaluator remains unbiased (for arbitrary reward estimator), and usually improves on IPS [8]. However, similar to IPS, it does not allow evaluation of nonstationary policies.
- Nonstationary Policy Evaluation [18, 19] uses rejection sampling (RS) to construct an unbiased history of interactions between the policy and the world. While this approach is unbiased, it may discard a large fraction of data through stringent rejection sampling, especially when the actions in the log are chosen from a highly non-uniform distribution. This can result in unacceptably large variance.

**Contributions.** In this paper, we propose a new policy evaluator that takes advantage of all good properties from the above approaches, while avoiding their drawbacks. As fundamental building blocks we use DR estimation, which

---

**Algorithm 1** DR-ns($\pi$, $\{(x_k, a_k, r_k, p_k)\}$, $q$, $c_{\max}$)

1. $h_0 \leftarrow \emptyset, t \leftarrow 1, c_1 \leftarrow c_{\max}$
   $R \leftarrow 0, C \leftarrow 0, Q \leftarrow \emptyset$
2. For $k = 1, 2, \ldots$ consider event $(x_k, a_k, r_k, p_k)$
   (a) $R_k \leftarrow \sum_{a'} \pi(a'|x_k, h_{t-1})\hat{r}(x_k, a')$
   $\qquad + \frac{\pi(a_k|x_k, h_{t-1})}{p_k} \cdot (r_k - \hat{r}(x_k, a_k))$
   (b) $R \leftarrow R + c_t R_k$
   (c) $C \leftarrow C + c_t$
   (d) $Q \leftarrow Q \cup \left\{\frac{p_k}{\pi(a_k|x_k, h_{t-1})}\right\}$
   (e) Let $u_k \sim \text{Uniform}[0,1]$
   (f) If $u_k \leq \frac{c_t \pi(a_k|x_k, h_{t-1})}{p_k}$
       i. $h_t \leftarrow h_{t-1} + (x_k, a_k, r_k)$
       ii. $t \leftarrow t + 1$
       iii. $c_t \leftarrow \min\{c_{\max}, q\text{-th quantile of } Q\}$
3. Return $R/C$

---

is extremely efficacious in stationary settings, and rejection sampling, which tackles nonstationarity. We introduce two additional strategies for variance control:

- In DR, we harness the knowledge of the randomness in the evaluated policy ("revealed randomness"). Randomization is the preferred tool for handling the exploration/exploitation tradeoff and if not properly incorporated into DR, it would yield an increase in the variance. We avoid this increase without impacting bias.
- We substatially improve sample use (i.e., acceptance rate) in rejection sampling by modestly increasing the bias. Our approach allows an easy control of the bias/variance tradeoff.

As a result, we obtain an evaluator of nonstationary policies, which is extremely sample-efficient while taking advantage of reward estimators and scavenged exploration through incorporation into DR. Our incorporation of revealed randomness yields a favorable bonus: when the past data is generated by the same (or very similar) policy as the one evaluated, we accept all (or almost all) samples—a property called "idempotent self-evaluation."

After introducing our approach in Sec. 2, we analyze its bias and variance in Sec. 3, and finally present an extensive empirical evaluation in Sec. 4.

## 2 A New Policy Evaluator

Algorithm 1 describes our new policy evaluator DR-ns (for "doubly robust nonstationary"). Over the run of the algorithm, we process the past deployment data (exploration samples) and run rejection sampling (Steps 2e–2f) to create a simulated history $h_t$ of the interaction between the target policy and the environment. The algorithm returns the expected reward estimate $R/C$.

The algorithm takes as input a target policy to evaluate, exploration samples, and two scalars $q$ and $c_{\max}$ which control the tradeoff between the length of $h_t$ and bias. On each exploration sample, we use a modified DR to estimate $\mathbf{E}_\pi[r_t|x_t = x_k, h_{t-1}]$ (Step 2a). Compared with Eq. (1.1), we take advantage of revealed randomness.

The rate of acceptance in rejection sampling is controlled by the variable $c_t$ that depends on two parameters: $c_{\max}$, controlling the maximum allowed acceptance rate, and $q$, which allows adaptive (policy-specific) adjustment of the acceptance rate. The meaning of $q$ is motivated by unbiased estimation as follows: to obtain no bias, the value of $c_t$ should never exceed the ratio $p_k/\pi(a_k|x_k, h_{t-1})$ (i.e., the right-hand side in Step 2f should never exceed one). During the run of the algorithm we keep track of the observed ratios (in $Q$), and $q$ determines the quantile of the empirical distribution in $Q$, which we use as an upper bound for $c_t$. Setting $q = 0$, we obtain the unbiased case (in the limit). By using larger values of $q$, we increase the bias, but get longer sequences through increased acceptance rate. Similar effect is obtained by varying the value of $c_{\max}$, but the control is cruder, since it ignores the evaluated policy.

In reward estimation, we weigh the current DR estimate by the current acceptance rate, $c_t$ (Step 2b). In unbiased case, for each simulated step $t$, we expect to accumulate multiple samples with the total weight of 1 in expectation. To obtain better scaling (similar to importance weighted methods), we also accumulate the sum of weights $c_t$ in the variable $C$, and use them to renormalize the final estimate as $R/C$.

As a quick observation, if we let $c_{\max} = 1$ and evaluate a (possibly randomized nonstationary) policy on data that this policy generated, every event is accepted into the history regardless of $q$. Note that such aggressive all-accept strategy is unbiased for this specific "self-evaluation" setting and makes the maximal use of data. Outside self-evaluation, $q$ has an effect on acceptance rate (stronger if exploration and target policy are more different). In our experiments, we set $c_{\max} = 1$ and rely on $q$ to control the acceptance rate.

## 3 Analysis

We start with basic definitions and then proceed to analysis.

### 3.1 Definitions and Notation

Let $D(x)$ and $D(r|x, a)$ denote the unknown (conditional) distributions over contexts and over rewards. To simplify notation (and avoid delving into measure theory), we assume that rewards and contexts are taken from some countable sets, but our theory extends to arbitrary measurable context spaces $\mathcal{X}$ and arbitrary measurable rewards in $[0, 1]$. We assume that actions are chosen from a finite set $\mathcal{A}$ (this is a critical assumption). Our algorithm also uses an estimator of expected conditional reward $\hat{r}(x, a)$, but we do not require that this estimator be accurate. For example, one can define $\hat{r}(x, a)$ as a constant function for some value in $[0, 1]$; often the constant may be chosen 0.5 as the minimax optimum [4]. However, if $\hat{r}(x, a) \approx \mathbf{E}_D[r|x, a]$, then our value estimator will have a lower variance but unchanged bias [8]. In our analysis, we assume that $\hat{r}$ is fixed and determined before we see the data (e.g., by initially splitting the input dataset).

We assume that the input data is generated by some past (possibly nonstationary) policy, which we refer to as the "exploration policy." Contexts, actions, and rewards observed by the exploration policy are indexed by timesteps $k = 1, 2, \ldots$. The input data consists of tuples $(x_k, a_k, r_k, p_k)$, where contexts and rewards are sampled according to $D$, and $p_k$ is the logged probability with which the action $a_k$ was chosen. In particular, we will *not* need to evaluate probabilities of choosing actions $a' \neq a_k$, nor require the full knowledge of the past policy, substantially reducing logging efforts.

Our algorithm augments tuples $(x_k, a_k, r_k, p_k)$ by independent samples $u_k$ from the uniform distribution over $[0, 1]$. A history up to the $k$-th step is denoted

$$z_k = (x_1, a_1, r_1, p_1, u_1, \ldots, x_k, a_k, r_k, p_k, u_k) \ ,$$

and an infinite history $(x_k, a_k, r_k, p_k, u_k)_{k=1}^\infty$ is denoted $z$. In our analysis, we view histories $z$ as samples from a distribution $\mu$. Our assumptions about data generation then translate into the assumption about factoring of $\mu$ as

$$\begin{aligned}\mu(x_k, a_k, & r_k, p_k, u_k|z_{k-1}) \\ &= D(x_k)\mu(a_k|x_k, z_{k-1})D(r_k|x_k, a_k) \\ &\quad I\left(p_k = \mu(a_k|x_k, z_{k-1})\right) U(u_k)\end{aligned}$$

where $U$ is the uniform distribution over $[0, 1]$. Note that apart from the unknown distribution $D$, the only degree of freedom above is $\mu(a_k|x_k, z_{k-1})$, i.e., the unknown exploration policy.

When $z_{k-1}$ is clear from the context, we use a shorthand $\mu_k$ for the distribution over the $k$-th tuple

$$\begin{aligned}\mu_k(x, a, & r, p, u) \\ & = \mu\left(x_k = x, \ a_k = a, \ r_k = r, \ p_k = p, \ u_k = u \ \big| \ z_{k-1}\right) \ .\end{aligned}$$

We also write $\mathbf{P}_k^\mu$ and $\mathbf{E}_k^\mu$ for $\mathbf{P}_\mu[\cdot|z_{k-1}]$ and $\mathbf{E}_\mu[\cdot|z_{k-1}]$.

For the target policy $\pi$, we index contexts, actions, and rewards by $t$. Finite histories of this policy are denoted as

$$h_t = (x_1, a_1, r_1, \ldots, x_t, a_t, r_t)$$

and the infinite history is denoted $h$. Nonstationary policies depend on a history as well as the current context, and hence can be viewed as describing conditional probability distributions $\pi(a_t|x_t, h_{t-1})$ for $t = 1, 2, \ldots$. In our analysis, we extend the nonstationary target policy $\pi$ into a probability distribution over $h$ defined by the factoring

$$\pi(x_t, a_t, r_t|h_{t-1}) = D(x_t)\pi(a_t|x_t, h_{t-1})D(r_t|x_t, a_t) \ .$$

Similarly to $\mu$, we define shorthands $\pi_t(x, a, r)$, $\mathbf{P}_t^\pi$, $\mathbf{E}_t^\pi$.

We assume a continuous running of our algorithm on an infinite history $z$. For $t \geq 1$, let $\kappa(t)$ be the index of the $t$-th sample accepted in Step 2f; thus, $\kappa$ converts an index in the target history into an index in the exploration history. We set $\kappa(0) = 0$ and define $\kappa(t) = \infty$ if fewer than $t$ samples are accepted. Note that $\kappa$ is a deterministic function of the history $z$. For simplicity, we assume that for every $t$, $\mathbf{P}_\mu[\kappa(t) = \infty] = 0$. This means that the algorithm generates a distribution over histories $h$, we denote this distribution $\hat\pi$.

Let $B(t) = \{\kappa(t-1) + 1, \kappa(t-1) + 2, \ldots, \kappa(t)\}$ for $t \geq 1$ denote the set of sample indices between the $(t-1)$-st acceptance and the $t$-th acceptance. This set of samples is called the $t$-th block. The inverse operator identifying the block of the $k$-th sample is $\tau(k) = t$ such that $k \in B(t)$. The contribution of the $t$-th block to the value estimator is denoted $R_{B(t)} = \sum_{k \in B(t)} R_k$. In our analysis, we assume a completion of $T$ blocks, and consider both normalized and unnormalized estimators:

$$R = \sum_{t=1}^T c_t R_{B(t)} \ , \quad \tilde R^{\text{avg}} = \frac{\sum_{t=1}^T c_t R_{B(t)}}{\sum_{t=1}^T c_t |B(t)|} \ .$$

### 3.2 Bias Analysis

Our goal is to develop an accurate estimator. Ideally, we would like to bound the error as a function of an increasing number of exploration samples. For nonstationary policy, it can be easily shown that a single simulation trace of a policy can yield reward that is bounded away by 0.5 from the expected reward regardless of the length of simulation (see, e.g., Example 3 in [19]). Hence, even for unbiased methods, we cannot accurately estimate the expected reward from a single trace.

A simple (but wasteful) approach is to divide the exploration samples into several parts, run the algorithm separately on each part, obtaining estimates $R^{(1)}, \ldots, R^{(m)}$, and return the average $\sum_{i=1}^m R^{(i)}/m$. (We only consider the unnormalized estimator in this section. We assume that the division into parts is done sequentially, so that each estimate is based on the same number of blocks $T$.) Using standard concentration inequalities, we can then show that the average is within $O(1/\sqrt{m})$ of the expectation $\mathbf{E}_\mu[R]$. The remaining piece is then bounding the bias term $\mathbf{E}_\mu[R] - \mathbf{E}_\pi[\sum_{t=1}^T r_t]$.

Recall that $R = \sum_{t=1}^T c_t R_{B(t)}$. The source of bias are events when $c_t$ is not small enough to guarantee that $c_t \pi(a_k|x_k, h_{t-1})/p_k$ is a probability. In this case, the probability that the $k$-th exploration sample is accepted is

$$p_k \min\left\{1, \tfrac{c_t\pi(a_k|x_k, h_{t-1})}{p_k}\right\} = \min\{p_k, c_t\pi(a_k|x_k, h_{t-1})\} \ ,$$

which violates the unbiasedness requirement that the probability of acceptance be proportional to $\pi(a_k|x_k, h_{t-1})$.

Let $\mathcal{E}_k$ denote this "bad" event (conditioned on $z_{k-1}$ and the induced target history $h_{t-1}$):

$$\mathcal{E}_k = \{(x,a) : c_t\pi_t(a|x) > \mu_k(a|x)\} \ .$$

Associated with this event is the "bias mass" $\varepsilon_k$:

$$\varepsilon_k = \mathop{\mathbf{P}}_{(x,a)\sim\pi_t}[\mathcal{E}_k] - \mathop{\mathbf{P}}_{(x,a)\sim\mu_k}[\mathcal{E}_k]/c_t \ .$$

Notice that from the definition of $\mathcal{E}_k$, this mass is non-negative. Since the first term is a probability, this mass is at most 1. We assume that this mass is bounded away from 1, i.e., that there exists $\varepsilon$ such that for all $k$ and $z_{k-1}$ we have the bound $0 \leq \varepsilon_k \leq \varepsilon < 1$.

The following theorem analyzes how much bias is introduced in the worst case, as a function of $\varepsilon$. Its purpose is to identify the key quantities that contribute to the bias, and to provide insights of what to optimize in practice.

**Theorem 1.** *For $T \geq 1$,*

$$\left|\mathbf{E}_\mu\left[\sum_{t=1}^T c_t R_{B(t)}\right] - \mathbf{E}_\pi\left[\sum_{t=1}^T r_t\right]\right| \leq \frac{T(T+1)}{2} \cdot \frac{\varepsilon}{1-\varepsilon} \ .$$

Intuitively, this theorem says that if a bias of $\varepsilon$ is introduced in round $t$, its effect on the sum of rewards can be felt for $T - t$ rounds. Summing over rounds, we expect to get an $O(\varepsilon T^2)$ effect on the unnormalized bias in the worst case or equivalently a bias of $O(\varepsilon T)$ on the average reward. In general a very slight bias can result in a significantly better acceptance rate, and hence longer histories (or more replicates $R^{(i)}$).

This theorem is the first of this sort for policy evaluators, although the mechanics of proving correctness are related to the proofs for model-based reinforcement-learning agents in MDPs (e.g., [14]). A key difference here is that we depend on a context with unbounded complexity rather than a finite state space.

Before proving Theorem 1, we state two technical lemmas (for proofs see Appendix B). Recall that $\hat\pi$ denotes the distribution over target histories generated by our algorithm.

**Lemma 1.** *Let $t \geq 1$, $k \geq 1$ and let $z_{k-1}$ be such that $\kappa(t-1) = k-1$. Let $h_{t-1}$ and $c_t$ be the target history and acceptance ratio induced by $z_{k-1}$. Then:*

$$\sum_{x,a}\left|\mathbf{P}_k^\mu[x_{\kappa(t)} = x, a_{\kappa(t)} = a] - \pi_t(x,a)\right| \leq \frac{2\varepsilon}{1-\varepsilon} \ ,$$

$$\left|c_t\mathbf{E}_k^\mu[R_{B(t)}] - \mathbf{E}_t^\pi[r_t]\right| \leq \frac{\varepsilon}{1-\varepsilon} \ .$$

**Lemma 2.** $\sum_{h_T} |\hat{\pi}(h_T) - \pi(h_T)| \leq (2\varepsilon T) / (1 - \varepsilon)$ .

*Proof of Theorem 1.* We first bound a single term $|\mathbf{E}_{z\sim\mu}[c_t R_{B(t)}] - \mathbf{E}_{h\sim\pi}[r_t]|$ using the previous two lemmas, the triangle inequality and Hölder's inequality:

$$\begin{aligned}
&\left|\mathbf{E}_{z\sim\mu}[c_t R_{B(t)}] - \mathbf{E}_{h\sim\pi}[r_t]\right| \\
&= \left|\mathbf{E}_{z\sim\mu}\left[c_t \mathbf{E}^\mu_{\kappa(t)}[R_{B(t)}]\right] - \mathbf{E}_{h\sim\pi}[r_t]\right| \\
&\leq \left|\mathbf{E}_{z\sim\mu}\left[\mathbf{E}^\pi_t[r_t]\right] - \mathbf{E}_{h\sim\pi}\left[\mathbf{E}^\pi_t[r_t]\right]\right| + \frac{\varepsilon}{1-\varepsilon} \\
&= \left|\mathbf{E}_{h_{t-1}\sim\hat{\pi}}\left[\mathbf{E}^\pi_t\left[r - \frac{1}{2}\right]\right] - \mathbf{E}_{h_{t-1}\sim\pi}\left[\mathbf{E}^\pi_t\left[r - \frac{1}{2}\right]\right]\right| + \frac{\varepsilon}{1-\varepsilon} \\
&\leq \frac{1}{2} \sum_{h_{t-1}} |\hat{\pi}(h_{t-1}) - \pi(h_{t-1})| + \frac{\varepsilon}{1-\varepsilon} \\
&\leq \frac{1}{2} \cdot \frac{2\varepsilon(t-1)}{1-\varepsilon} + \frac{\varepsilon}{1-\varepsilon} = \frac{\varepsilon t}{1-\varepsilon} \ .
\end{aligned}$$

The theorem now follows by summing over $t$ and using the triangle inequality. □

### 3.3 Progressive Validation

While the bias analysis in the previous section qualitatively captures the bias-variance tradeoff, it cannot be used to construct an explicit error bound. The second and perhaps more severe problem is that even if we had access to a more explicit bias bound, in order to obtain *deviation* bounds, we would need to decrease the length of generated histories by a significant factor (at least according to the simple approach discussed at the beginning of the previous section).

In this section, we show how we can use a single run of our algorithm to construct a stationary policy, whose value is estimated with an error $O(1/\sqrt{n})$ where $n$ is the number of original exploration samples. Thus, in this case we get explicit error bound and much better sample efficiency.

Assume that the algorithm terminates after fully generating $T$ blocks. We will show that the value $R/C$ returned by our algorithm is an unbiased estimate of the expected reward of the randomized stationary policy $\pi_{\text{PV}}$ defined by:

$$\pi_{\text{PV}}(a|x) = \sum_{t=1}^T \frac{c_t |B(t)|}{C} \pi(a|x, h_{t-1}) \ .$$

Conceptually, this policy first picks among the histories $h_0, \ldots, h_{T-1}$ with probabilities $c_1|B(1)|/C, \ldots, c_t|B(T)|/C$, and then executes the policy $\pi$ given the chosen history. We extend $\pi_{\text{PV}}$ to a distribution over triples

$$\pi_{\text{PV}}(x, a, r) = D(x) \pi_{\text{PV}}(a|x) D(r|x, a) \ .$$

To analyze our estimator, we need to assume that during the run of the algorithm, the ratio $\pi_t(a_k|x_k)/\mu_k(a_k|x_k)$ is bounded, i.e., we assume there exists $M < \infty$ such that

$$\forall z, \forall t \geq 1, \forall k \in B(t): \quad \frac{\pi_t(a_k|x_k)}{\mu_k(a_k|x_k)} \leq M \ .$$

Next, fix $z_{k-1}$ and let $t = \tau(k)$ (note that $\tau(k)$ is a deterministic function of $z_{k-1}$). It is not too difficult to show that $R_k$ is an unbiased estimator of $\mathbf{E}_{r\sim\pi_t}[r]$, and to bound its range and variance (for proofs see Appendices B and C):

**Lemma 3.** $\mathbf{E}^\mu_k[R_k] = \mathbf{E}_{r\sim\pi_t}[r]$.

**Lemma 4.** $|R_k| \leq 1 + M$.

**Lemma 5.** $\mathbf{E}^\mu_k[R_k^2] \leq 3 + M$.

Now we are ready to show that $R/C$ converges to the expected reward of the policy $\pi_{\text{PV}}$:

**Theorem 2.** *Let $n$ be the number of exploration samples used to generate $T$ blocks, i.e., $n = \sum_{t=1}^T |B(t)|$. With probability at least $1 - \delta$,*

$$\left|R/C - \mathbf{E}_{r\sim\pi_{\text{PV}}}[r]\right| \leq \frac{nc_{\max}}{C} \cdot 2 \max\left\{\frac{(1+M)\ln(2/\delta)}{n}, \sqrt{\frac{(3+M)\ln(2/\delta)}{n}}\right\} \ .$$

*Proof.* The proof follows by Freedman's inequality (Theorem 3 in Appendix A), applied to random variables $c_t R_k$, whose range and variance can be bounded using Lemmas 4 and 5 and the bound $c_t \leq c_{\max}$. □

## 4 Experiments

We conduct experiments on two problems, the first is a public supervised learning dataset converted into an exploration learning dataset, and the second is a real-world proprietary dataset.

### 4.1 Classification with Bandit Feedback

In the first set of experiments, we illustrate the benefits of DR-ns (Algorithm 1) over naive rejection sampling using the public dataset rcv1 [17]. Since rcv1 is a multi-label dataset, an example has the form $(x, c)$, where $x$ is the feature and $c$ is the set of corresponding labels. Following the construction of previous work [4, 8], an example $(x, c)$ in a $K$-class classification problem may be interpreted as a bandit event with context $x$, action $a \in [K] := \{1, \ldots, K\}$, and loss $l_a := I(a \notin c)$, and a classifier as an arm-selection policy whose expected loss is its classification error. In this section, we aim at evaluating average policy loss, which can be understood as negative reward. For our experiments, we only use the $K = 4$ top-level classes in rcv1, namely $\{C, E, G, M\}$; a random selection of $40K$ data from the whole dataset were used. Call this dataset $D$.

**Data Conversion.** To construct a partially labeled exploration data set, we choose actions non-uniformly in the following manner. For an example $(x, c)$, a uniformly random

score $s_a \in [0.1, 1]$ is assigned to arm $a \in [K]$, and the probability of action $a$ is

$$\mu(a|x) = \frac{0.3 \times s_a}{\sum_{a'} s_{a'}} + \frac{0.7 \times I(a \in c)}{|c|}.$$

This kind of sampling ensures two useful properties. First, every action has a non-zero probability, so such a dataset suffices to provide an unbiased offline evaluation of any policy. Second, actions corresponding to correct labels have higher observation probabilities, emulating the typical setting where a baseline system already has a good understanding of which actions are likely best.

We now consider the two tasks of evaluating a static policy and an adaptive policy. The first serves as a sanity check to see how well the evaluator works in the degenerate case of static policies. In each task, a one-vs-all reduction is used to induce a multi-class classifier from either fully or partially labeled data. In fully labeled data, each example is included in data sets of all base binary classifiers. In partially labeled data, an example is included only in the data set corresponding to the action chosen. We use the LIB-LINEAR [9] implementation of logistic regression. Given a classifier that predicts the most likely label $a^*$, our policy follows an $\varepsilon$-greedy strategy with $\varepsilon$ fixed to $0.1$; that is, with probability $0.9$ it chooses $a^*$, otherwise a random label $a \in [K]$. The reward estimator $\hat{r}$ is directly obtained from the probabilistic output of LIBLINEAR (using the "-b 1" option). The scaling parameter is fixed to the default value 1 (namely, "-c 1").

**Static Policy Evaluation.** In this task, we first chose a random $10\%$ of $D$ and trained a policy $\pi_0$ on this fully labeled data. From the remainder, we picked a random "evaluation set" containing $50\%$ of $D$. The average loss of $\pi_0$ on the evaluation set served as the ground truth. A partially labeled version of the evaluation set was generated by the conversion described above; call the resulting dataset $D'$. Finally, various offline evaluators of $\pi_0$ were compared against each other on $D'$.[1] We repeated the generation of evaluation set in 300 trials to measure the bias and standard deviation of each evaluator.

**Adaptive Policy Evaluation.** In this task, we wanted to evaluate the average online loss of the following adaptive policy $\pi$. The policy is initialized as a specific "offline" policy calculated on random 400 fully observed examples ($1\%$ of $D$). Then the "online" partial-feedback phase starts (the one which we are interested in evaluating). We update the policy after every 15 examples, until 300 examples are observed. On policy update, we simply use an enlarged training set containing the initial 400 fully labeled and additional partially labeled examples. The offline training set was fixed in all trials. The remaining data was split into two portions, the first containing a random $80\%$ of $D$ for evaluation, the second containing $19\%$ of $D$ to determine the ground truth. The evaluation set was randomly permuted and then transformed into a partially labeled set $D'$ on which evaluators were compared. The generation of $D'$ was repeated in 50 trials, from which bias and standard deviation of each evaluator were obtained. To estimate the ground-truth value of $\pi$, we simulated $\pi$ on the randomly shuffled (fully labeled) $19\%$ of ground-truth data $2\,000$ times to compute its average online loss.

---

[1] To avoid risks of overfitting, for evaluators that estimate $\hat{r}$ (all except RS), we split $D'$ into two equal halves, one for training $\hat{r}$, the other for running the evaluator. The same approach was taken in the adaptive policy evaluation task.

**Compared Evaluators.** We compared the following evaluators described earlier: DM for direct method, RS for the unbiased evaluator in [19] combined with rejection sampling, and DR-ns as in Algorithm 1 (with $c_{\max} = 1$). We also tested a variant of DR-ns, which does not monitor the quantile, but instead uses $c_t$ equal to $\min_D \mu(a|x)$; we call it WC since it uses the worst-case (most conservative) value of $c_t$ that ensures unbiasedness of rejection sampling.

**Results.** Tables 1 and 2 summarize the accuracy of different evaluators in the two tasks, including rmse (root mean squared error), bias (the absolute difference between evaluation mean and the ground truth), and stdev (standard deviation of the evaluation results in different runs). It should be noted that, given the relatively small number of trials, the measurement of bias is not statistically significant. So for instance, it cannot be inferred in a statistically significant way from Table 1 that WC enjoys a lower bias than RS. However, the tables provide $95\%$ confidence interval for the rmse that allows a meaningful comparison.

It is clear from both tables that although rejection sampling is guaranteed to be unbiased, its variance usually is the dominating source of rmse. At the other extreme is the direct method, which has the smallest variance but often suffers high bias. In contrast, our method DR-ns is able to find a good balance between these extremes and, with proper choice of $q$, is able to yield much more accurate evaluation results. Furthermore, compared to the unbiased variant WC, DR-ns's bias appears to be modest.

It is also clear that the main benefit of DR-ns is its low variance, which stems from the adaptive choice of $c_t$ values. By slightly violating the unbiasedness guarantee, it increases the effective data size significantly, hence reducing the variance of its evaluation. In particular, in the first task of evaluating a static policy, rejection sampling was able to use only 264 examples (out of the 20K data in $D'$) since the minimum value of $\mu(a|x)$ in the exploration data was very small; in contrast, DR-ns was able to use 523, 3 375, 4 279, and 4 375 examples for $q \in \{0, 0.01, 0.05, 0.1\}$, respectively. Similarly, in the adaptive policy evaluation task, with DR-ns($q > 0$), we could extract many more online tra-jectories of length 300 for evaluating $\pi$, while RS and WC were able to find only one such trajectory out of the evaluation set. In fact, if we increased the trajectory length of $\pi$ from 300 to 500, neither RS or WC could construct a full trajectory of length 500 and failed the task completely.

Table 1: Static policy evaluation results.

| evaluator | rmse ($\pm 95\%$ C.I.) | bias | stdev |
|---|---|---|---|
| DM | $0.0151 \pm 0.0002$ | 0.0150 | 0.0017 |
| RS | $0.0191 \pm 0.0021$ | 0.0032 | 0.0189 |
| WC | $\mathbf{0.0055 \pm 0.0006}$ | 0.0001 | 0.0055 |
| DR-ns($q=0$) | $0.0093 \pm 0.0010$ | 0.0032 | 0.0189 |
| DR-ns($q=0.01$) | $0.0057 \pm 0.0006$ | 0.0021 | 0.0053 |
| DR-ns($q=0.05$) | $\mathbf{0.0055 \pm 0.0006}$ | 0.0022 | 0.0051 |
| DR-ns($q=0.1$) | $0.0058 \pm 0.0006$ | 0.0017 | 0.0055 |

Table 2: Adaptive policy evaluation results.

| evaluator | rmse ($\pm 95\%$ C.I.) | bias | stdev |
|---|---|---|---|
| DM | $0.0329 \pm 0.0007$ | 0.0328 | 0.0027 |
| RS | $0.0179 \pm 0.0050$ | 0.0007 | 0.0181 |
| WC | $0.0156 \pm 0.0037$ | 0.0086 | 0.0132 |
| DR-ns($q=0$) | $0.0129 \pm 0.0034$ | 0.0046 | 0.0122 |
| DR-ns($q=0.01$) | $\mathbf{0.0089 \pm 0.0017}$ | 0.0065 | 0.0062 |
| DR-ns($q=0.05$) | $0.0123 \pm 0.0017$ | 0.0107 | 0.0061 |
| DR-ns($q=0.1$) | $0.0946 \pm 0.0015$ | 0.0946 | 0.0053 |

### 4.2 Content Slotting in Response to User Queries

In this set of experiments, we evaluate two policies on a proprietary real-world dataset consisting of web search queries, various content that is displayed on the web page in response to these queries, and the feedback that we get from the user (as measured by clicks) in response to the presentation of this content. Formally, this partially labeled data consists of tuples $(x_k, a_k, r_k, p_k)$, where $x_k$ is a query and corresponding features, $a_k \in \{\text{web-link,news,movie}\}$ (the content shown at slot 1 on the results page), $r_k$ is a so-called click-skip reward ($+1$ if the result was clicked, $-1$ if a result at a *lower* slot was clicked), and $p_k$ is the recorded probability with which the exploration policy chose the given action.

The page views corresponding to these tuples represent a small percentage of traffic for a major website; any given page view had a small chance of being part of this experimental bucket. Data was collected over a span of several days during July 2011. It consists of 1.2 million tuples, out of which the first 1 million were used for estimating $\hat{r}$ with the remainder used for policy evaluation. For estimating the variance of the compared methods, the latter set was divided into 10 independent test subsets of equal size.

Two policies were compared in this setting: *argmax* and *self-evaluation* of the exploration policy. For *argmax* pol-icy, we first obtained a linear estimator $r'(x,a) = w_a \cdot x$ by importance-weighted linear regression (with importance weights $1/p_k$). The *argmax* policy chooses the action with the largest predicted reward $r'(x,a)$. Note that both $\hat{r}$ and $r'$ are linear estimators obtained from the training set, but $\hat{r}$ was computed without importance weights (and we therefore expect it to be more biased). *Self-evaluation* of the exploration policy was performed by simply executing the exploration policy on the evaluation data.

Table 3: Estimated rewards reported by different policy evaluators on two policies for a real-world exploration problem. In the first column results are normalized by the (known) expected reward of the deployed policy. In the second column results are normalized by the reward reported by IPS. All $\pm$ are computed standard deviations over results on 10 disjoint test sets.

| evaluator | *self-evaluation* | *argmax* policy |
|---|---|---|
| RS | $0.986 \pm 0.060$ | $0.990 \pm 0.048$ |
| IPS | $0.995 \pm 0.041$ | $1.000 \pm 0.027$ |
| DM | $1.213 \pm 0.010$ | $1.211 \pm 0.002$ |
| DR | $0.967 \pm 0.042$ | $0.991 \pm 0.026$ |
| DR-ns | $0.974 \pm 0.039$ | $0.993 \pm 0.024$ |

Table 3 compares RS [19], IPS, DM, DR [8], and DR-ns($c_{\max} = 1$, $q = 0.1$). For business reasons, we do not report the estimated reward directly, but normalize to either the empirical average reward (for *self-evaluation*) or the IPS estimate (for the *argmax* policy evaluation).

In both cases, the RS estimate has a much larger variance than the other estimators. Note that the minimum observed $p_k$ equals $1/13$, which indicates that a naive rejection sampling approach would suffer from the data efficiency problem. Indeed, out of approximately 20 000 samples per evaluation subset, about 900 are added to the history for the *argmax* policy. In contrast, the DR-ns method adds about 13 000 samples, a factor of 14 improvement.

The experimental results are generally in line with theory. The variance is smallest for DR-ns, although IPS does surprisingly well on this data, presumably because values $\hat{r}$ in DR and DR-ns are relatively close to zero, so the benefit of $\hat{r}$ is diminished. The Direct Method (DM) has an unsurprisingly huge bias, while DR and DR-ns appear to have a very slight bias, which we believe may be due to imperfect logging. In any case, DR-ns dominates RS in terms of variance as it was designed to do, and has smaller bias and variance than DR.

## 5 Conclusion and Future Work

We have unified best-performing stationary policy evaluators and rejection sampling by carefully preserving their best parts and eliminating the drawbacks. To our knowl-edge, the resulting approach yields the best evaluation method for nonstationary and randomized policies, especially when reward predictors are available.

Yet, there are definitely opportunities for further improvement. For example, consider nonstationary policies which can devolve into a round-robin action choices when the rewards are constant (such as UCB1 [2]). A policy which cycles through actions has an expected reward equivalent to a randomized policy which picks actions uniformly at random. However, for such a policy, our policy evaluator will only accept on average a fraction of $1/K$ uniform random exploration events. An open problem is to build a more data-efficient policy evaluator for this kind of situations.